\documentclass[sigconf]{acmart}

\usepackage{amsfonts}
\usepackage{amsmath}
\usepackage{graphicx} 
\usepackage{array,color,booktabs,multirow,boldline,makecell}
\usepackage{balance}
\copyrightyear{2020}
\acmYear{2020}
\setcopyright{acmcopyright}\acmConference[CIKM '20]{Proceedings of the 29th ACM International Conference on Information and Knowledge Management}{October 19--23, 2020}{Virtual Event, Ireland}
\acmBooktitle{Proceedings of the 29th ACM International Conference on Information and Knowledge Management (CIKM '20), October 19--23, 2020, Virtual Event, Ireland}
\acmPrice{15.00}
\acmDOI{10.1145/3340531.3411904}
\acmISBN{978-1-4503-6859-9/20/10}

\begin{document}
\fancyhead{}

\title{Opinion-aware Answer Generation for Review-driven Question Answering in E-Commerce}\thanks{The work described in this paper is substantially supported by a grant from the Research Grant Council of the Hong Kong Special Administrative Region, China (Project Codes: 14200719).}

\author{Yang Deng, Wenxuan Zhang, Wai Lam}
\affiliation{%
  \institution{The Chinese University of Hong Kong}
}
\email{{ydeng, wxzhang, wlam}@se.cuhk.edu.hk}

\begin{abstract}
Product-related question answering (QA) is an important but challenging task in E-Commerce. It leads to a great demand on automatic review-driven QA, which aims at providing instant responses towards user-posted questions based on diverse product reviews. 
Nevertheless, the rich information about personal opinions in product reviews, which is essential to answer those product-specific questions, is underutilized in current generation-based review-driven QA studies. 
There are two main challenges when exploiting the opinion information from the reviews to facilitate the opinion-aware answer generation: (i) jointly modeling opinionated and interrelated information between the question and reviews to capture important information for answer generation, (ii) aggregating diverse opinion information to uncover the common opinion towards the given question.
In this paper, we tackle opinion-aware answer generation by jointly learning answer generation and opinion mining tasks with a unified model. Two kinds of opinion fusion strategies, namely, static and dynamic fusion, are proposed to distill and aggregate important opinion information learned from the opinion mining task into the answer generation process. Then a multi-view pointer-generator network is employed to generate opinion-aware answers for a given product-related question. Experimental results show that our method achieves superior performance in real-world E-Commerce QA datasets, and effectively generate opinionated and informative answers.
\end{abstract}


\maketitle

\section{Introduction}\label{intro}
Product-related question answering (QA), which aims at solving product-specific questions, has drawn extensive attention due to its broad application in real-world E-Commerce sites, such as Amazon\footnote{http://www.amazon.com} and eBay\footnote{http://www.ebay.com}.
These E-Commerce sites are usually equipped with a community question answering (CQA) system, enabling users to address their concerns by interacting with other users through questions and answers. However, there exist a large number of questions posted but unanswered. In such case, the user reviews, which contain the user personal opinions and actual experiences about the concerned product, can be utilized to provide responses. 
This leads to the rapid development of review-driven QA for automatically providing answers via analysis of product reviews~\cite{DBLP:conf/www/McAuleyY16,DBLP:conf/icdm/WanM16,DBLP:conf/aaai/0007GZZWZS19,DBLP:conf/wsdm/ChenLJZC19}, which can be served as an AI assistant helping manage the tremendous amount of product reviews and provide a possible solution to those unanswered questions. 

\begin{table*}
\caption{Two examples from Amazon QA platform with relevant review snippets.}
\centering
  \begin{tabular}{cp{6.8cm}p{7.3cm}}
  \toprule
  \textbf{Question}& Is this really a good buy for cycling?& Does this device play Blu-ray on PC? \\
  \midrule
  \textbf{Reference Answer} & Yes it is.  It's pretty well padded. & Yes. It comes with software to play Blu-ray discs on a PC. \\
  \midrule
  \multirow{3}{*}{\textbf{\shortstack{Relevant\\Review\\Snippets\\(Partial)}}}& 1. This seat is comfortable and works well. I mentioned to my sister, about wanting to buy a new seat, and she suggested buying a gel cover.& 1. No complaints one way or the other, the device works as expected and allowed me to view Blu-ray disks on a PC that didn't have a Blu-ray device. \\
  & 2. This cushion is high quality and very comfy. This one holds up even if it gets wet.& 2. I have only used it on a XBMC and Win 7 PC for movies since I don't own a player for Mac.\\
  & 3. Not perfect in my opinion, could have used a bit more padding. & 3. Again, it does not play discs, so what did I just buy?\\
  \midrule
  \midrule
  \textbf{Approach}&\multicolumn{2}{c}{\textbf{Answer}}\\
  \midrule
  \textbf{Opinion-based}& Yes& Yes\\
  \midrule
  \textbf{Retrieval-based}& This cushion is high quality and very comfy. This one holds up even if it gets wet. & No complaints one way or the other, the device works as expected and allowed me to view Blu-ray disks on a PC that didn't have a Blu-ray device. \\
  \midrule
  \textbf{Generation-based}& I don't think so. I don't think it would be too big for cycling.& I don't see why it wouldn't work with the Blu-ray player, but it does have an HDMI input.  \\
  \midrule
  \midrule
  \multirow{1}{*}{\textbf{\shortstack{Opinion-aware\\Generation}}}& Yes, it is a very good seat. I have been using it for several months now and have not had any problems.& Yes, it does work with the Blu-ray player. I haven’t had any problems with them at all.\\
  \bottomrule
  \end{tabular}
\vspace{-0.3cm}
\label{example}
\end{table*}

\citet{DBLP:conf/www/McAuleyY16} report an observation that about 56\% of questions in Amazon are explicitly asking for certain opinion. 
This indicates the necessity of considering opinions from former buyers in automatic product-related question answering. The notion of ``opinion” here is essentially a generalized form of what it refers to in traditional opinion mining tasks. It can be interpreted as the belief or stance held by the users towards the given question.  
Such questions include both subjective questions asking about positive/negative feeling or stance, and objective questions confirming the actual product details, as shown in the first and second questions in Table~\ref{example}, respectively. 
From another perspective, users who post the question would also like to know about others' actual opinions towards the question, either positive or negative, rather than some randomly generated opinions.
Meanwhile, product reviews preserve a wide range of both objective and subjective product-related information. Beyond telling us subjective opinions that whether a product is ``good" or ``bad", which is the main goal of traditional opinion mining or sentiment analysis in review datasets~\cite{DBLP:conf/coling/KimH04,DBLP:conf/semeval/PontikiGPMA15}, reviews also provide a wide range of actual experiences, including objective descriptions of products’ properties, functional assessments, specific use-cases and so on.
Thus, it is of great importance to take into account customers' opinions reflected in the reviews when providing answers for product-related questions.

Early review-driven QA studies  typically adopt two kinds of approaches, namely opinion-based and retrieval-based. Opinion-based approaches aim to predict ``yes/no" answers based on the opinions in relevant reviews~\cite{DBLP:conf/www/McAuleyY16,DBLP:conf/icdm/WanM16}, while retrieval-based approaches retrieve the most related review snippet as the answer~\cite{DBLP:conf/emnlp/YuZC12,DBLP:conf/aaai/0007GZZWZS19}. 
Recently, inspired by the successful applications in machine translation~\cite{DBLP:conf/nips/SutskeverVL14} and summarization~\cite{DBLP:conf/acl/SeeLM17}, text generation methods are proposed to generate natural sentences as the answer from relevant product reviews~\cite{DBLP:conf/wsdm/ChenLJZC19,DBLP:conf/wsdm/GaoRZZYY19}. 
Two real-world examples from Amazon are presented in Table~\ref{example}, which provide the generated answers by current review-driven QA approaches.
It can be observed that opinion-based approach only gives the classification result of the answer type, based on the common opinion reflected in the product reviews, without detailed information. 
Retrieval-based approach selects the most related review as the answer, which cannot answer the given question precisely since the review is not specifically written for answering the given question. 
While providing natural forms of answers, there are some defects on the generated answers by current generation-based approaches. The answers provided by this kind of approaches often hold a random opinion towards the given question, even contradictory to the common opinion among the relevant reviews, as the examples presented in Table~\ref{example}. The reason is that current generation-based approaches indifferently take into account all the relevant reviews with diverse opinions towards the given question, neglecting the opinion information reflected in the review, which is shown to be crucial in product-related QA problem~\cite{DBLP:conf/www/McAuleyY16,DBLP:conf/icdm/WanM16}. Therefore, in this work, we study opinion-aware answer generation for review-driven QA, which aims at generating natural answers that are aware of customers' opinions from the reviews for product-specific questions.


There are two challenges for  incorporating opinion information into review-driven answer generation: (1) The reviews of the same question may differ in customers' opinions, which makes it difficult to aggregate the opinion information into the final generated answer. For instance, back to the first example in Table~\ref{example}, there are two relevant reviews (\#1 \& \#2) holding a positive opinion corresponding to the given question, but the third review claims a relatively negative opinion. Similarly, in the second example, there also exists opinion divergence in the customers' actual experience towards such an objective question. 
(2) The opinion information and the interactions between question and reviews are supposed to possess mutual inference in determining the importance of each review on answer generation. Intuitively, the opinion information in the most relevant review to the given question is supposed to be more important in determining the opinion type in the generated question. On the other hand, the decisive reviews in mining the common opinion among all the relevant reviews are supposed to be more influential in generating opinion-aware answers.

To tackle these issues, we aim to generate opinion-aware natural answers via multi-task learning~\cite{DBLP:conf/ijcai/MaSLR18,DBLP:conf/aaai/DengXLYDFLS19} to conduct answer generation and opinion mining tasks simultaneously. Specifically, we first adopt a co-attentive matching layer to capture the relevant information between the question and reviews. Then we conduct opinion mining to identify the core opinion of those relevant reviews towards the given question, as well as fetch common opinion information for answer generation. Finally, a multi-view pointer-generator network is exploited to combine the important information from both the question and reviews. We further propose two kinds of opinion fusion mechanism, static and dynamic fusion, to refine and incorporate opinion information for generating opinion-aware answers.

In summary, our main contributions are as follows:

(1) We exploit opinion information reflected in the reviews, which is underutilized in existing works, for review-driven answer generation.

(2) We tackle this problem by jointly learning answer generation and opinion mining tasks with a unified model.

(3) We propose a multi-view pointer-generator network with static and dynamic opinion fusion to integrate information from different perspectives for opinion-aware answer generation.

(4) Our method outperforms existing methods on real-world E-Commerce QA datasets and effectively generates opinionated and informative answers.

\section{Related Work}
\subsection{Review-driven Question Answering}
Different from general community question answering~\cite{DBLP:conf/semeval/NakovMMMGR15}, which aims at ranking a set of answer candidates, most of studies on product-related question answering in E-Commerce scenario exploit product reviews to provide the answer. 
According to different strategies to provide answers, existing review-driven QA methods can be categorized into four groups. (1) \textbf{Opinion-based  methods}~\cite{DBLP:conf/www/McAuleyY16,DBLP:conf/icdm/WanM16,DBLP:conf/wsdm/YuL18} conduct a classification task to provide yes/no answers by identifying customers' common opinion type from the relevant reviews towards the question. For example, \citet{DBLP:conf/www/McAuleyY16} construct an opinion question answering dataset from Amazon QA platform, which aims at classifying the answer opinion type by mining relevant opinions from reviews. (2) \textbf{Retrieval-based methods}~\cite{DBLP:conf/emnlp/YuZC12,DBLP:conf/aaai/0007GZZWZS19} aim to retrieve review sentences as the answer by ranking the relevance between the question and reviews. Besides, some studies~\cite{DBLP:conf/wsdm/YuQJHSCC18,DBLP:conf/sigir/ZhangDL20} follow the traditional CQA problem setting to rank a list of user-written answers or predict their helpfulness~\cite{DBLP:conf/www/ZhangLDM20}, instead of using product reviews. (3) \textbf{Query-based summarization methods}~\cite{DBLP:conf/coling/WangRCC14,DBLP:conf/sigir/LiuFPHY16} summarize the review sentences as the answer with the guidance of question information. (4) \textbf{Text generation methods}~\cite{DBLP:conf/wsdm/ChenLJZC19,DBLP:conf/wsdm/GaoRZZYY19} adopt seq2seq based neural networks to generate fluent sentences as answers. \citet{DBLP:conf/wsdm/ChenLJZC19} exploit both the attention and gate mechanism to capture the relevant information from the reviews to alleviate the noise issue in review-driven answer generation. \citet{DBLP:conf/wsdm/GaoRZZYY19} incorporate product attribute to extract helpful facts for generating answers from reviews. In this work, we study opinion-aware answer generation to generate more meaningful and helpful answers for product-related questions.

\subsection{Opinion Mining \& Sentiment Analysis}
Traditional opinion mining and sentiment analysis  studies in E-Commerce scenario mainly focus on sentiment classification~\cite{DBLP:conf/coling/KimH04,DBLP:conf/semeval/PontikiGPMA15} or rating prediction~\cite{DBLP:conf/sigir/JinLZZLWM16,DBLP:conf/www/ChengDZK18}. Some latest studies jointly learn sentiment analysis with other tasks. \citet{DBLP:conf/emnlp/ShenSWKLLSZZ18} propose a novel problem, QA-style sentiment classification, aiming at addressing sentiment analysis in QA applications, which is further studied with reinforcement learning by \citet{DBLP:conf/acl/WangSLLSZZ19}.
Besides, sentiment-aware review summarization recently gains increasingly attention. \citet{DBLP:conf/coling/YangQSLZZ18} and \citet{DBLP:conf/cikm/TianY019} extract aspect and sentiment words or lexicons to facilitate the sentiment-aware review summarization. 
\citet{DBLP:conf/ijcai/MaSLR18} and \citet{DBLP:conf/acml/WangR18} exploit multi-task learning methods to conduct sentiment classification in product reviews with the text summarization as an auxiliary task. To the best of our knowledge, this work is the first attempt to jointly learn opinion mining and answer generation tasks.

\subsection{Text Generation}
Recent years have witnessed many successful applications of sequence-to-sequence~\cite{DBLP:conf/nips/SutskeverVL14} based model on text generation tasks. Most of existing generation methods are developed by employing attention mechanism~\cite{DBLP:journals/corr/BahdanauCB14} and pointer-generator network~\cite{DBLP:conf/acl/SeeLM17}. Some latest studies attempt to generate target text from multi-document or multi-passage source text. \citet{DBLP:conf/acl/SunHLLMT18} and \citet{DBLP:conf/acl/NishidaSNSOAT19} jointly learn sentence extraction and text generation. Other related works leverage various external information to enrich the generated text. \citet{DBLP:conf/cikm/SunJSPOW18} and \citet{DBLP:journals/corr/abs-1909-02745} incorporate external knowledge to enhance the generation performance. Apart from review-driven answer generation~\cite{DBLP:conf/wsdm/ChenLJZC19,DBLP:conf/wsdm/GaoRZZYY19}, recently, generative question answering has also been explored in reading comprehension~\cite{DBLP:conf/acl/NishidaSNSOAT19,DBLP:journals/corr/abs-1909-02745} and community question answering scenario~\cite{DBLP:journals/corr/abs-1911-09801,DBLP:journals/corr/abs-1912-00864}. In this work, the reviews are regarded as a kind of multi-passage external sources for improving the product-related answer generation.

\section{Problem Definition}
Given a question $q$ and a series of relevant reviews $\{r_1,...,r_K\}$, the goal is to simultaneously predict the common opinion polarity $l$, i.e., \textit{positive}, \textit{negative}, or \textit{neutral}, towards the given question and generate a natural language answer $y$ as the response. 

Specifically, the dataset $\mathbf{D}$ consists of $N$ data samples, in which the $i$-th data sample contains a question $q^i$, a set of auxiliary reviews $r^i$ with corresponding ratings $e^i$ to the product, a reference answer $a^i$ and an opinion type label $l^i$ of the answer. The dataset $\mathbf{D}$ can be represented by:
\begin{equation}
 \mathbf{D} = \{(q^i, \{(r^i_1,e^i_1),...,(r^i_K,e^i_K)\}, a^i, l^i\}^{N}_{i=1}.
\end{equation}

The goal is to generate the answer $y^i$ that can not only precisely answer the given question $q^i$ but also be coherent to the common opinion reflected in the relevant reviews $r^i$.

\section{Proposed Framework}
\begin{figure}
\centering
\includegraphics[width=0.48\textwidth]{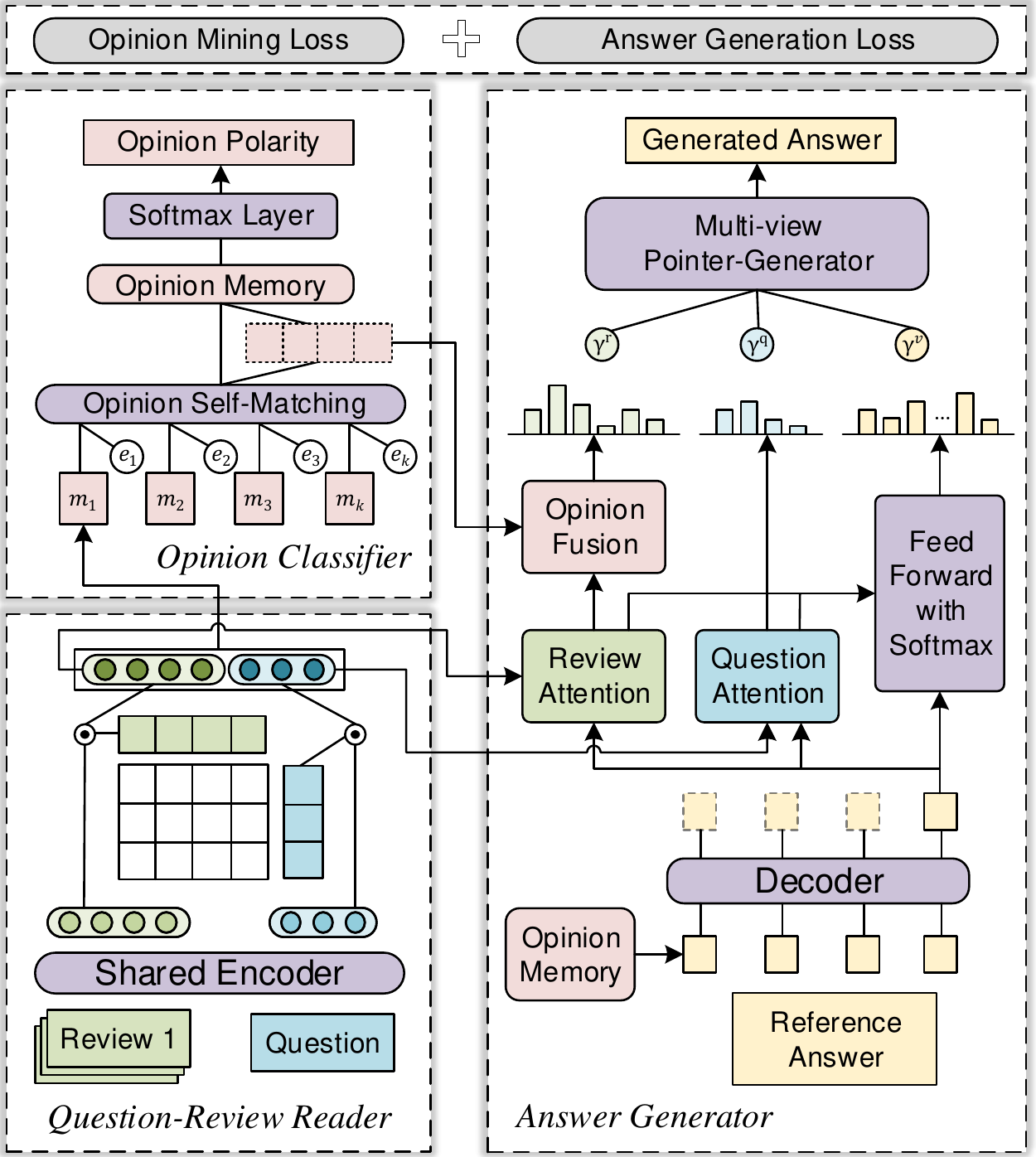}
\caption{Overview of Opinion-Aware Answer Generator}
\label{method}
\vspace{-0.2cm}
\end{figure}

We introduce Opinion-Aware Answer Generator (OAAG) to tackle opinion-aware answer generation by the multi-task learning of answer generation and opinion mining tasks.
Figure~\ref{method} depicts the overview of OAAG, which can be organized as three components: (1) \textit{Question-Review Reader} encodes the interrelation information between the question and each review into the sentence representations (Section~\ref{sec-reader}), (2) \textit{Opinion Classifier} extracts the opinion information from the related product reviews for the opinion-aware answer generation (Section~\ref{sec-classifier}), (3) \textit{Answer Generator} generates opinion-aware answers by taking into account both the interrelation information between the question and reviews and the opinion information among the reviews (Section~\ref{sec-generator}). The overall framework is trained on an end-to-end fashion under multi-task learning paradigm (Section~\ref{sec-training}).

\subsection{Question-Review Reader}\label{sec-reader}
Question-Review Reader aims to encode the raw text of questions and reviews into vector representations, by capturing the interrelation information between the question and each review.
\subsubsection{\textbf{Question-Review Encoder}}
At the beginning, each word in the question $q$ and $k$-th review $r_k$ is passed through an embedding layer. The word embeddings of the question and the review, $W_q$ and $W_{r_k}$, are fed into a Bi-LSTM encoder to learn both the head-to-tail and the tail-to-head context information:
\begin{gather}
H_{q}=\textbf{Bi-LSTM}(W_q),\quad
H_{r_k}=\textbf{Bi-LSTM}(W_{r_k}).
\end{gather}

We then encode word sequences of the question $q$ and the review $r_k$ into sentence representations $H_{q}, H_{r_k}\in\mathbb{R}^{L\times{d_h}}$, where $L$ and $d_h$ are the length of sentences and the size of hidden states respectively.

\subsubsection{\textbf{Co-Attentive Matching Layer}}
We apply a dual attention mechanism to compute the co-attention between the question representation $H_q$ and the $k$-th review representation $H_{r_k}$:
\begin{gather}
 \Omega_{q{r_k}}=\text{tanh}\left(H_qUH_{r_k}^\intercal\right),\\
 \alpha_{q_k} = \text{softmax}(\text{Max}(\Omega_{q{r_k}})),\\
 \alpha_{r_k} =\text{softmax}(\text{Max}(\Omega_{q{r_k}}^\intercal)),
\end{gather}
where $U\in\mathbb{R}^{d_{h}\times{d_{h}}}$ is the attention parameter matrix to be learned; $\text{Max}(\cdot)$ denotes row-wise max-pooling operation;  $\alpha_{q_k}$ and $\alpha_{r_k}$ are the co-attention weights between the question and the $k$-th review.

We conduct element-wise product, which is denoted by $\odot$, between the attention vectors and the question and review representations to generate the attentive representations. To obtain the final question representations, a mean-pooling operation is applied over the attentive question representations with all the reviews. As for the review, all the attentive representations for each review are sequentially concatenated to form the overall review representations: 
\begin{gather}
 \Pi_q = \frac{1}{K}\sum_{k=1}^K H_q \odot \alpha_{q_k} \\
 \Pi_r = [H_{r_1} \odot \alpha_{r_1};...;H_{r_K} \odot \alpha_{r_K}],
\end{gather}
where [;] denotes the sequential concatenation operation, $\Pi_q=\{\pi^q_1,...,\pi^q_{l_q}\}$ and $\Pi_r=\{\pi^r_1,...,\pi^r_{l_r}\}$ are the attentive encoded representations of questions and reviews, respectively.

Meanwhile, we concatenate the attentive question and review representations to form the matching vector $m_k$ as input of the opinion classifier for extracting the opinion information:
\begin{gather}
 m_k = [H_q^\intercal\alpha_{q_k} : H_{r_k}^\intercal\alpha_{r_k}],
\end{gather}
where [:] denotes the concatenation operation.

\subsection{Opinion Classifier}\label{sec-classifier}
Under the multi-task learning setting, the opinion classifier not only detects the opinion towards the product-related question, but also guides the answer generator to be aware of important opinion information from product reviews.

\subsubsection{\textbf{Opinion Self-Matching Layer}}
After encoding sentences into vector representations with the Question-Review Reader, the attentive matching vectors $m_k$ are generated to pinpoint the interrelated information in each question-review pair.

Intuitively, the degree of importance and relatedness is supposed to be diverse in different question-review pairs. Moreover, those related reviews may contain different opinions towards the concerned question due to different individual user experience. Therefore, we design a self-matching layer to aggregate the related and influential reviews for inferring the general and common opinion among the reviews towards the given question and differentiating the value of each review. 

Since every review comes with a rating (e.g., 1 to 5) given by the same user, which also reflects the user's subjective opinion towards the product, we concatenate each matching vector $m_k$ with the corresponding one-hot rating embedding $e_k$,
$\hat{m}_k = [m_k:e_k]$. We calculate the review-level attention weights with the final matching vectors $\hat{m}_k$ by the following vanilla attention mechanism:
\begin{gather}
    M = [\hat{m}_1;\hat{m}_2;...;\hat{m}_K],\\
    U_m = \text{tanh}(W_m M),\\
    \beta = \text{softmax}(\omega_m^\intercal U_m),
\end{gather}
where $\beta$ is the review-level opinion attention weight which measures the importance degree of each review in determining the common opinion among all the reviews, $W_m$ and $\omega_m$ are the attention matrices to be learned. 

Then we derive the final opinion memory representation by the dot product of the matching vectors and the opinion attention weights:
\begin{equation}
    \hat{O} = M^\intercal\beta.
\end{equation}

\subsubsection{\textbf{Opinion Classification}}
The opinion memory representation $\hat{O}$, which contains the core opinion information reflected in the relevant review set, then is fed into a softmax layer for the opinion classification:
\begin{equation}
p^{o}=\text{softmax}(W_s\hat{O}+b_s),
\end{equation}
where $p^{o}$ is the predicted probability of the answer opinion polarities, i.e., \textit{positive}, \textit{negative} and \textit{neutral}.  $W_s\in\mathbb{R}^{d_m\times{3}}$ and $b_s\in\mathbb{R}^3$ are the trainable weight matrix and bias vector in the hidden layer.

\subsection{Answer Generator}\label{sec-generator}
Answer generator integrates the opinion information learned from the opinion classifier and the relevant information learned from the question-review reader to generate natural language answers for the given question.

\subsubsection{\textbf{Attention-based Decoder}}
We adopt a unidirectional LSTM as the decoder. The opinion representation $\hat{O}$ is exploited as the initial decoder state $s_0$, which enables the decoder to begin decoding with certain opinion information. At each step $t$, the decoder produces hidden state $s_t$ with the input of the previous word $w_{t-1}$:
\begin{equation}
    s_t= \textbf{LSTM}(s_{t-1},w_{t-1}).
\end{equation}

The attention weight for each word in the question and the review, $\alpha^q_t$ and $\alpha^r_t$, are generated by:
\begin{gather}
e^{q_j}_t = \omega_q^\intercal \text{tanh} (W_{q} \pi^q_j+W_{qs} s_t+b_{q}), \\
\alpha^q_t = \text{softmax}(e^q_t),\\
e^{r_i}_t = \omega_r^\intercal \text{tanh} (W_{r} \pi^r_i+W_{rs} s_t+b_{r}), \\
\alpha^r_t = \text{softmax}(e^r_t),
\end{gather}
where $W_{q}$, $W_{qs}$, $W_{r}$, $W_{rs}$, $\omega_q$, $\omega_r$, $b_{q}$, $b_{r}$ are parameters to be learned. The attention weights $\alpha^q_t$ and $\alpha^r_t$ are used to compute context vectors $c^q_t$ and $c^r_t$ as the probability distribution over the source words:
\begin{gather}
    c^q_t = \sum\nolimits_j^{l_q} \alpha^{q_j}_t \pi^q_j, \quad
    c^r_t = \sum\nolimits_i^{l_r} \alpha^{r_i}_t \pi^r_i.
\end{gather}

The context vector aggregates the information from the source text for the current step. We concatenate the context vector with the decoder state $s_t$ and pass it through a linear layer to generate the answer representation $h^s_t$:
\begin{equation}
    h^s_t=W_1[s_t:c^q_t:c^r_t]+b_1,
\end{equation}
where $W_1$ and $b_1$ are parameters to be learned.

\subsubsection{\textbf{Opinion Fusion}}
In addition to the basic pointer-generator network, the model copies words not only from the question but also from the reviews. In order to attend words in reviews with decisive opinion and also alleviate the noise from irrelevant reviews, we introduce two strategies of \textit{opinion fusion} to re-weight attention scores of the words in reviews. 

\noindent \textbf{Static Fusion.}
The word attention of reviews are combined with the static review-level attention weights $\beta$ learned from the opinion mining task, which measure the importance of each review in determining the answer opinion polarity. Thus, the opinion fusion function is defined as:
\begin{equation}
    \hat{\alpha}^{r_i}_t = \frac{\alpha^{r_i}_t \beta_{r_i \in l}}{\sum_i \alpha^{r_i}_t  \beta_{r_i \in l}}.
\end{equation}

Note that different from existing attention combination method~\cite{DBLP:conf/acl/SunHLLMT18}, static fusion combines the word and review level attentions from two different perspectives.

\noindent \textbf{Dynamic Fusion.}
In static fusion, the diversity of the generated answer will be limited, since the review-level attention weight $\beta$ remains unchanged during the decoding procedure in the same case. Therefore, we propose dynamic fusion to address this issue. The attentive opinion matching vectors, $o_k=\beta_k\hat{m_k}$, are leveraged to dynamically generate review-level attention weights along with the decoding procedure:
\begin{gather}
    e^{o_k}_t = \omega_o^\intercal \text{tanh} (W_{o} o_k+W_{os} s_t+b_{o}), \\
    \hat{\beta}_t = \text{softmax}(e^o_t),
\end{gather}
where $W_{o}$, $W_{os}$, $\omega_o$, $b_{o}$ are parameters to be learned. Thus the dynamic review attention $\hat{\beta}_t$ will replace $\beta$ in Eq.19 to compute the dynamic fusion for each decoding step $t$.

Thus, the re-weighted word-level attention weights in reviews will be:
\begin{equation}
    \hat{\alpha}^{r_i}_t = \frac{\alpha^{r_i}_t \hat{\beta}_{t,{r_i \in l}}}{\sum_i \alpha^{r_i}_t  \hat{\beta}_{t,{r_i \in l}}}.
\end{equation}

\subsubsection{\textbf{Multi-view Pointer-Generator}}
A multi-view pointer-generator architecture with opinion fusion strategy is designed to generate opinion-aware answers as well as handle the out-of-vocabulary (OOV) issue. Such approach makes the decoder capable to copy words from the question and be aware of opinion words from reviews. 

First, the probability distribution $P^v$ over the fixed vocabulary is obtained by passing the answer representation $h^s_t$ through a softmax layer:
\begin{equation}
    P^v(y_t) = \text{softmax}(W_2 h^s_t + b_2),
\end{equation}
where $W_2$ and $b_2$ are parameters to be learned. 

Then, we obtain the attention-based probability distribution by indexing the attention weight of each word in both the question and review to the extended vocabulary:
\begin{gather}
    P^q(y_t) = \sum\nolimits_{i:w_i=w} \alpha^{q_i}_t,\quad
    P^r(y_t) = \sum\nolimits_{i:w_i=w} \hat{\alpha}^{r_i}_t,
\end{gather}
where $P^q(y_t)$ and $P^r(y_t)$ denote the attention-based probability distribution for the question words and the review words, respectively.

The final probability distribution of $y_t$ is obtained from three views of word distributions, including the question attention based, the review attention based and the original vocabulary probability distribution.
\begin{gather}
    P^{all}(y_t) = [P^v(y_t),P^q(y_t),P^r(y_t)],\\
    \gamma = \text{softmax}(W_{\gamma}[s_t:c^q_t:c^r_t] + b_{\gamma}),\\
    P(y_t) = \sum \gamma P^{all}(y_t),
\end{gather}
where $W_\gamma$ and $b_\gamma$ are parameters to be learned, $\gamma$ is the multi-view pointer scalar to determine the weight of each view of probability distribution.

\subsection{Multi-Task Learning Procedure}\label{sec-training}
Finally, we conduct multi-task learning for the proposed framework, which jointly learn the opinion mining and the answer generation tasks by an end-to-end training procedure.
\subsubsection{\textbf{Opinion Mining Loss}}
The opinion mining task is trained to minimize the cross-entropy loss function:
\begin{equation}
 L_{om}=-\sum\nolimits_{i=1}^Nl_i\log{p^o_i},
\end{equation}
where $p^o$ is the output of the opinion classifier and $l$ is the opinion type label of the answer.

\subsubsection{\textbf{Answer Generation Loss}}
The answer generation task is trained to minimize the negative log likelihood:
\begin{equation}
    L_{ag} = - \frac{1}{T}\sum\nolimits^T_{t=0}\text{log}P(w_t^*).
\end{equation}

\subsubsection{\textbf{Overall Loss Function}}
For joint training, the final objective function is to minimize the overall loss function:
\begin{equation}
    L=L_{om}+\lambda L_{ag},
\end{equation}
where $\lambda$ is a hyper-parameter to balance losses.

\section{Experiments}
\subsection{Research Questions}
The empirical analysis targets at the following research questions: 
\begin{itemize}
    \item \textbf{RQ1}: What is the overall performance of OAAG? Does it outperform state-of-the-art baselines? 
    \item \textbf{RQ2}: How does each component in OAAG contribute to the overall performance?
    \item \textbf{RQ3}: How does OAAG address the concerning issues discussed in Section~\ref{intro}?
    \item \textbf{RQ4}: How does OAAG perform when generating answers with different kinds of opinions? \item \textbf{RQ5}: What is the difference in the generated answers given by the two variants of OAAG, i.e., using static opinion fusion and dynamic opinion fusion? 
\end{itemize}

\subsection{Dataset}
We evaluate our model on Amazon Question Answering Dataset~\cite{DBLP:conf/www/McAuleyY16}, 
which contains around 1.4 million answered questions with answer opinion label, including \textit{positive}, \textit{negative}, and \textit{neutral}, from different categories. 
This QA dataset can be combined with Amazon Product Review Dataset~\cite{DBLP:conf/www/HeM16}, by matching the product ID. Since there are a large number of reviews for each product, we need to extract those reviews that contain relevant information for each question. 
Similar to~\citet{DBLP:conf/wsdm/ChenLJZC19}, each review text is chunked into  snippets of length 50, or to the end of a sentence boundary. Then for a given question, we adopt BM25 to rank all the review snippets of the corresponding product and collect top 10 relevant review snippets for each question as the model input.

After we collect the final dataset, each QA sample contains a question, a reference answer, the answer opinion type label, and a set of relevant review snippets with corresponding ratings. Three categories with the largest number of samples are adopted, namely \textit{Electronics}, \textit{Home\&Kitchen} and \textit{Sports\&Outdoors}. For each category, we split 10\% instances for evaluation, and the remaining are used for training. The statistics of the dataset\footnote{https://github.com/dengyang17/OAAG} are presented in Table~\ref{dataset}.

\begin{table}
\centering
\caption{The statistic of datasets}
  \begin{tabular}{c|c||ccc}
  \toprule
  Dataset & Set & \#(Q,A) & Avg QLen & Avg ALen\\
  \midrule
  \midrule
  \multirow{2}{*}{Electronics} & Train&174,565&16.37&39.63\\
   & Test & 19,395&16.71&37.34\\
  \midrule
  Home\& &Train&81,250&15.38&35.85\\
  Kitchen & Test &9,019&15.64&35.90\\
  \midrule
  Sports\&&Train&45,018&15.61&35.66\\
  Outdoors & Test&5,002&16.04&37.21\\
  \bottomrule
  \end{tabular}
\label{dataset}
\end{table}

\begin{table*}[htb]
\centering
\caption{\label{ansgen} Method comparisons and ablation studies on answer generation task}
\begin{tabular}{lccccccccccccccc}
\toprule
\multirow{2}{*}{Model}& \multicolumn{5}{c}{Electronics}& \multicolumn{5}{c}{Home\&Kitchen}&\multicolumn{5}{c}{Sports\&Outdoors} \\ 
\cmidrule(lr){2-6}\cmidrule(lr){7-11}\cmidrule(lr){12-16}
&R1&RL&B1&ES&TOA&R1&RL&B1&ES&TOA&R1&RL&B1&ES&TOA\\
\midrule
BM25& 9.6&8.3&6.0 &81.0&50.8&9.0&8.0& 5.4 &80.1&56.6&9.5&8.2&5.9&80.7&54.3\\
\midrule
S2SA~\cite{DBLP:journals/corr/BahdanauCB14} &14.4&12.9&12.5&85.2&53.4&14.1&12.9&12.4&85.4&56.9& 13.3&12.3&12.1&84.2&57.8 \\
PGN~\cite{DBLP:conf/acl/SeeLM17} & 11.3&10.1&10.5&83.5&51.2&11.6&10.2&10.7&83.9&56.1& 10.6&9.3&10.0&81.9&55.4\\
S2SAR~\cite{DBLP:conf/wsdm/GaoRZZYY19} &14.7 &13.1&12.6&85.4&53.6&\underline{14.9}&\underline{13.7}&\underline{12.9}&84.4&55.7&13.5&12.5&12.2&84.6&55.2\\
QS~\cite{DBLP:journals/corr/abs-1712-06100} & 13.5&12.5&11.6&84.6&53.5& 14.0&12.9&12.3&83.8&62.2& 14.3&13.2&12.2&84.1&53.3 \\
RAGE~\cite{DBLP:conf/wsdm/ChenLJZC19} &14.5 &12.9&\underline{12.7}&85.3&56.3&\underline{14.9}&13.4&12.4&\underline{85.7}&60.8&14.9&13.7&13.1&85.0&58.5\\
\midrule
HSSC-Q~\cite{DBLP:conf/ijcai/MaSLR18} & 14.3 &13.3&12.4&85.3& \underline{56.6}&14.3&13.1&12.8&85.2&62.8&15.0&13.8&13.0&84.6&59.3\\
SAHSSC-Q~\cite{DBLP:conf/acml/WangR18} &\underline{14.8} &\underline{13.6}&12.6&\underline{85.6}& 54.3&13.9&12.9&12.6&84.4&\underline{63.4}&\underline{15.3}&\underline{14.1}&\underline{13.3}&\underline{85.2}&\underline{60.2}\\
\midrule
\midrule
\textbf{OAAG-S}&15.8&\textbf{14.5}&13.4&85.4&60.3&\textbf{16.6}&\textbf{15.1}&\textbf{14.3}&86.0&68.5& 16.0&\textbf{14.5}&\textbf{13.6}&\textbf{85.9}&65.0\\
\textbf{OAAG-D} &\textbf{15.9}&\textbf{14.5}&\textbf{13.5}&\textbf{85.7}&\textbf{61.8}&16.4&\textbf{15.1}&13.9&\textbf{86.3}&\textbf{69.9}&\textbf{16.1}&14.3&\textbf{13.6}&85.4&\textbf{66.8}\\
- co-attentive&15.0&14.1&13.1&85.4&60.1&15.9&14.7&13.6&85.7&65.4&15.8&14.2&13.3&85.3&63.7\\
- opinion memory & 15.4&14.3&13.3&85.3&58.8&15.9&14.7&13.7&85.5&62.7&15.7&14.0&13.3&85.4&61.1\\
- opinion fusion & 15.3&14.0&13.0&85.3&59.2&15.6&14.3&13.4&84.9&64.3&15.4&14.0&13.2&85.5&63.3\\

\bottomrule
\end{tabular}
\end{table*}

\subsection{Baseline Methods \& Evaluation Metrics}
OAAG-S denotes our proposed model, OAAG, with the static opinion fusion strategy, while OAAG-D denotes OAAG with dynamic fusion. We compare with several baselines and state-of-the-art methods on both answer generation task and opinion mining task as well as some related multi-task learning models.  Following the previous works on review-driven answer generation~\cite{DBLP:conf/wsdm/ChenLJZC19,DBLP:conf/wsdm/GaoRZZYY19}, we adopt five generation-based methods for answer generation task:
\begin{itemize}
    \item \textbf{S2SA}~\cite{DBLP:journals/corr/BahdanauCB14}. The standard Seq2Seq model with attention mechanism. The input sequence is only the question.
    \item \textbf{PGN}~\cite{DBLP:conf/acl/SeeLM17}. An abstractive summarization model copies words from the reviews with a pointer network, and produces new words by an encoder-decoder network. PGN generates answers from reviews without the question.
    \item \textbf{S2SAR}~\cite{DBLP:conf/wsdm/GaoRZZYY19}. A method incorporates the review information into S2SA model, by concatenating the question and all the reviews as the source text.
    \item \textbf{QS}~\cite{DBLP:journals/corr/abs-1712-06100}. A query-based summarization model regards product reviews as the original article, the question as a query and it generates the summary as the answer.
    \item \textbf{RAGE}~\cite{DBLP:conf/wsdm/ChenLJZC19}. A state-of-the-art review-driven answer generation framework for product-related questions\footnote{https://github.com/WHUIR/RAGE}. For a fair comparison, RAGE/POS is adopted, which is a variant of RAGE model without exploiting the POS tag features.
\end{itemize}

Four sentiment analysis models are adopted for the comparison of the opinion mining task: 
\begin{itemize}
    \item \textbf{Bi-LSTM}. A standard bidirectional LSTM model which concatenates the question and review text as a sequence for sentiment classification.
    \item \textbf{IAN}~\cite{DBLP:conf/ijcai/MaLZW17}. An approach considers both attention mechanisms on the aspect and the full context. 
    \item \textbf{MGAN}~\cite{DBLP:conf/emnlp/FanFZ18}. A fine-grained attention mechanism to model the interaction between the aspect and its context on the word-level. As for our implementations of IAN and MGAN, we adopt the question as the aspect information for the aspect-based sentiment analysis methods\footnote{https://github.com/songyouwei/ABSA-PyTorch}.
    \item \textbf{HMN}~\cite{DBLP:conf/emnlp/ShenSWKLLSZZ18}. A hierarchical matching network for QA-style sentiment classification. We regard the review as the answer text for opinion classification.
\end{itemize}

In addition, we adapt two multi-task learning
models of abstractive summarization and sentiment classification to the defined multi-task setting: 
\begin{itemize}
    \item \textbf{HSSC-Q}. HSSC~\cite{DBLP:conf/ijcai/MaSLR18} is an unified model jointly learns summarization and sentiment classification tasks\footnote{https://github.com/lancopku/HSSC}. We implement a simple method, HSSC-Q, which can incorporate the question information for the joint learning of answer generation and opinion mining tasks. Specifically, we concatenate the question and reviews as the input of the model.
    \item \textbf{SAHSSC-Q}. SAHSSC~\cite{DBLP:conf/acml/WangR18} is a self-attentive hierarchical model for jointly improving text summarization and sentiment classification. Same as HSSC-Q, we encode the question and the reviews into the joint learning model.
\end{itemize}  

We adopt ROUGE F1 (R1, RL), BLEU (B1) and Embedding-based Similarity (ES)~\cite{DBLP:conf/emnlp/LiuLSNCP16} as evaluation metrics to measure the performance of answer generation, and also adopt human evaluation and Distinct scores for analysis. In addition, similar to some text style transfer studies~\cite{DBLP:conf/nips/ShenLBJ17,DBLP:conf/acl/LiWZXRSZ18,DBLP:conf/acl/WuRLS19}, we train an opinion classifier with BERT~\cite{DBLP:conf/naacl/DevlinCLT19} on the reference answers and answer opinion types, then, the target opinion accuracy (TOA) of the generated answers is reported to evaluate the precision of the opinion type in the generated answers. MACRO-F1 and Accuracy are reported in the evaluation of opinion mining.

\subsection{Implementation Details}
We train all the implemented models with pre-trained GloVe embeddings\footnote{http://nlp.stanford.edu/data/glove.42B.zip} of 300 dimensions as word embeddings and set the vocabulary size to 50k. During training and testing procedure, we restrict the length of answers within 100 words. We train all the models for 20 epochs.  In our model, we train with a learning rate of 0.15 and an initial accumulator value of 0.1. The dropout rate is set to 0.5. The hidden unit sizes of the BiLSTM encoder and the LSTM decoder are all set to 256. We train our models with the batch size of 32. All other parameters are randomly initialized from [-0.05, 0.05]. $\lambda$ is set to 5.

\subsection{Results}
\subsubsection{\textbf{Answer Generation Results}}
For research question \textbf{RQ1}, which aims at demonstrating the effectiveness of OAAG, we evaluate the overall performance from diverse perspectives and compare with a variety of state-of-the-art methods. Answer generation results are summarized in Table~\ref{ansgen}, which shows that the proposed models, OAAG-S and OAAG-D, achieve the best performance in both content-preservation metrics (ROUGE, BLEU and Embedding-based Similarity) and opinion-accuracy metric (Target Opinion Accuracy) for the generated answers. 

There are several notable observations. (i) Methods that consider both question and review texts (S2SAR, QS, RAGE) perform better than basic generation methods (S2SA, PGN), indicating that it is necessary to take into account both the question and the review information when generating answers in E-Commerce scenario. (ii) A slight change of existing multi-task learning model for adaptation on answer generation task (HSSC-Q, SAHSSC-Q) shows not much improvement on the performance with other generation methods. We conjecture that the opinion information is not utilized to generate the answers in these multi-task learning models, so that opinion mining barely contributes to the answer generation. Besides, the interactions between the question and reviews are neglected in these two models, which also leads to the unsatisfied performance. (iii) OAAG substantially enhances the performance in all domains by carefully modeling the interactions between the question and reviews as well as the opinion information into answer generation process. Most importantly, we observe that OAAG makes a remarkable performance boosting, about 6\%, on the target opinion accuracy (TOA), implying that OAAG effectively and precisely generates opinion-aware answers. 

\subsubsection{\textbf{Ablation Study}}
For research question \textbf{RQ2}, we conduct several ablation studies to validate the effectiveness of certain components in the proposed model, and the results are presented in Table~\ref{ansgen}.  (i) Discarding co-attentive matching layer casts a negative impact on the answer generation performance, which demonstrates the importance of modeling the interactions between question and reviews. (ii) The ablation study in terms of discarding opinion memory or opinion fusion shows that incorporating opinion information actually improves the answer generation performance. Especially for the TOA metric, the performance suffers a large decrease when disabling opinion memory or opinion fusion module. This result indicates that the careful design for opinion mining actually assists in generating opinion-aware answers.

In addition, the ablation studies also validate the assumption that both the interrelated information and the opinionated information between the question and reviews should be taken into account for review-driven answer generation in E-Commerce scenario. This provides a partial answer to the research question \textbf{RQ3} that OAAG addresses the first issue for opinion-aware answer generation, concerning the joint modeling of interrelated and opinionated information between the question and reviews.

\subsubsection{\textbf{Human Evaluation}}
\begin{table}
\caption{Human evaluation results}
\centering
  \begin{tabular}{lccccc}
  \toprule
  Method & Info & Flu & Corr & Opn & Help\\
  \midrule
  \textsc{RAGE} & 3.38&3.49&3.61&3.23&3.54\\
  \textsc{SAHSSC-Q} &3.07& 3.54&3.37&3.52&3.23\\
  \textbf{OAAG-S} &3.63& 3.67& \textbf{3.89}&\textbf{4.05}&3.71 \\
  \textbf{OAAG-D} & \textbf{4.05}& \textbf{3.74}& 3.83&4.01&\textbf{3.92}\\
  \bottomrule
  \end{tabular}

\label{human_eval}
\vspace{-0.3cm}
\end{table}

We conduct human evaluation to evaluate the generated answers from five aspects: (1) Informativity: how rich is the generated answer in information? (2) Fluency: how fluent is the generated answer? (3) Correlatedness: how correlated is the generated answer to the given question? (4) Opinion: how well does the generated answer match the target opinion type? (5) Helpfulness: how helpful is the generated answer to the user?  We randomly sample 50 questions from each category and generate their answers with four methods, including RAGE, SAHSSC-Q and the proposed OAAG-S and OAAG-D. Three annotators are asked to score each generated answer with 1 to 5 (higher the better). 

The results in Table~\ref{human_eval} show that OAAG consistently outperforms other methods in five aspects. Noticeably, OAAG-D can provide richest information in generated answers, since it dynamically considers the information from different reviews. Besides, RAGE and OAAG generate answers more related to the question by taking into account the interactions between question and reviews. SAHSSC-Q and OAAG perform better in generating opinion-aware answers than RAGE. Overall, OAAG generates the most helpful answers with diverse, related, and opinionated information for users. The results further answer \textbf{RQ1} that the proposed method outperforms state-of-the-art methods from these practical perspectives.

\begin{table}
\centering
\caption{\label{omresult} Opinion mining results}
\begin{tabular}{p{1.48cm}cccccc}
\toprule
\multirow{2}{*}{Model}& \multicolumn{2}{c}{Electronics}&\multicolumn{2}{c}{Home}& \multicolumn{2}{c}{Sports} \\ 
\cmidrule(lr){2-3}\cmidrule(lr){4-5}\cmidrule(lr){6-7}
&F1&Acc&F1&Acc&F1&Acc\\
\midrule
BiLSTM &0.412&0.644&0.426&0.722&0.413&0.671\\
IAN &0.446&0.646&0.452&0.723&0.451&0.677\\
MGAN &0.466&0.646&0.460&0.725&0.444&0.674\\
HMN &\underline{0.471}&\underline{0.650}&\underline{0.486}&\underline{0.727}&\underline{0.464}&\underline{0.681}\\
\midrule
HSSC-Q &0.460&0.646&0.468&0.725&0.455&0.674 \\
SAHSSC-Q &0.465&0.648&0.470&0.723&0.463&0.675\\
\midrule
\midrule
\textbf{OAAG-S}&\textbf{0.491}&\textbf{0.656}&0.494&\textbf{0.732}&\textbf{0.493}&\textbf{0.684}\\
\textbf{OAAG-D}&0.481&0.654&\textbf{0.499}&0.731&0.490&0.680\\
- ratings&0.475&0.653&0.488&0.731&0.482&0.678\\
\bottomrule
\end{tabular}
\vspace{-0.3cm}
\end{table}

\subsubsection{\textbf{Opinion Mining Results}}
Opinion mining results are reported in Table~\ref{omresult}. Although opinion mining only serves as an auxiliary task of answer generation, OAAG also achieves competitive results with some state-of-the-art methods on opinion mining. 
Among the baseline methods, the QA-style sentiment classification method, HMN, achieves the best performance, while two multi-task learning methods with summarization module, HSSC-Q \& SAHSSC-Q, barely improve the performance from single-task methods. The results indicate the necessity of considering the interactions between the question and reviews when uncovering the common opinion towards the given question among all the reviews. 

Meanwhile, the strong performance on the opinion mining task guarantees the answer generation process to follow a precise guidance of opinions, which provides the other part of the answer to  \textbf{RQ3}, concerning the identification of the common opinion among diverse reviews. In addition, the performance can be benefited from adding the ratings of reviews into the opinion mining.

\subsection{Discussions}
\subsubsection{\textbf{Answers with Different Opinions}}
\begin{figure}
\centering
\includegraphics[width=0.48\textwidth]{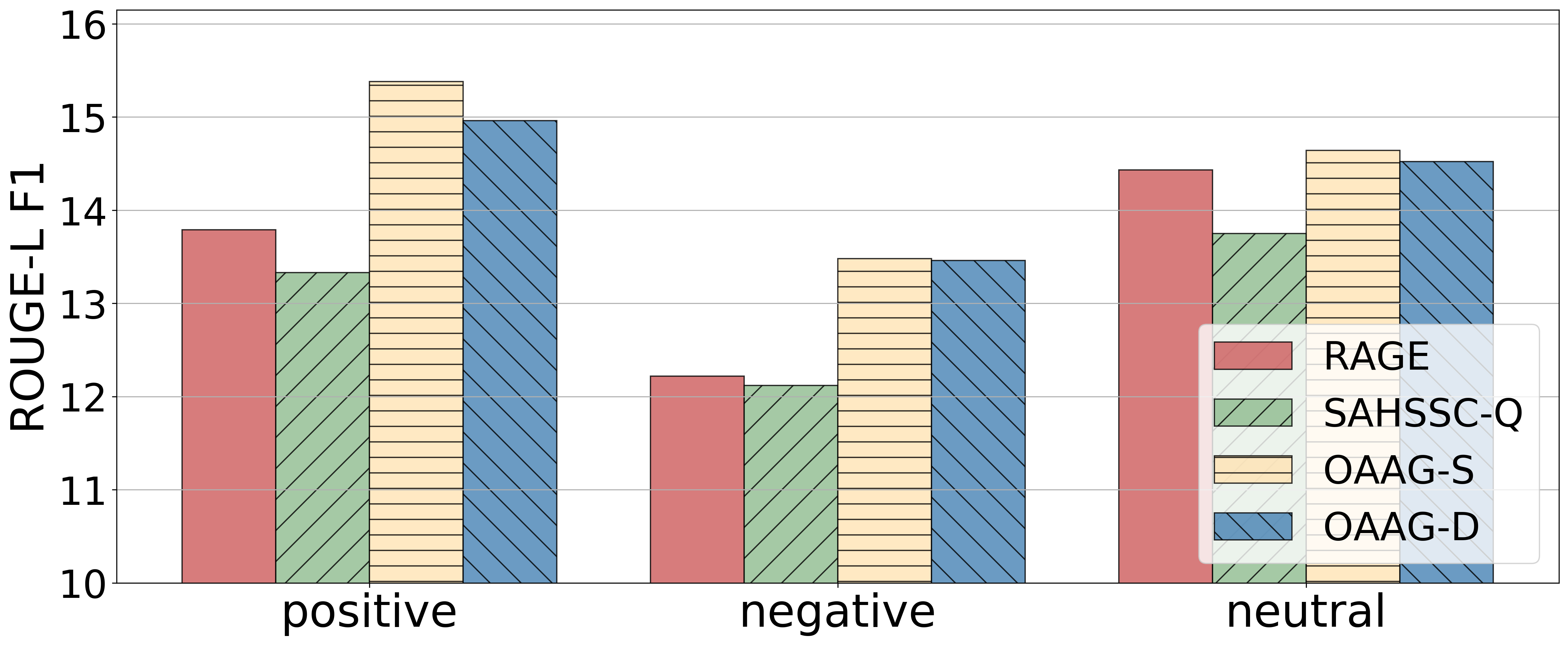}
\caption{Evaluation on answers in terms of opinions}
\label{type}
\vspace{-0.3cm}
\end{figure}

To address the research question \textbf{RQ4}, we evaluate the performance of OAAG on different answer types by ROUGE-L F1. As shown in Figure~\ref{type}, we observe that the proposed model outperforms the other two baselines (RAGE and SAHSSC-Q) on all types of answers. Worthy to note that generating precise answers with positive or negative opinions is more difficult than generating neutral answers, as it can be observed that the ROUGE scores of answers with positive or negative opinions are relatively lower than that of neutral answers for RAGE and SAHSSC-Q. However, OAAG shows a significant improvement on opinionated answer generation, which also demonstrates the effectiveness of incorporating opinion information. For those neutral answers, OAAG also maintains a high performance.

In addition, Table~\ref{trigram} reports the most frequent trigrams of the reference answer and generated answers given by different models in terms of opinion type. We observe that there is an obvious opinion polarity in reference answers, while the generated answers by RAGE miss the opinion information. Since SAHSSC-Q and OAAG take into account the opinion of reviews, the generated answers are distinguishable in different opinion types. Besides, it is interesting to see that RAGE and SAHSSC-Q tend to generate some meaningless neutral answers, such as ``i don't know" or ``i'm not sure", while OAAG alleviates this issue by highly interacting with reviews. 

Overall, OAAG can not only precisely generate answers with certain opinions, but also alleviate the issue of producing meaningless neutral answers. 

\begin{table}
\fontsize{8}{11}\selectfont
\centering
\caption{Trigram in different opinions}
  \begin{tabular}{|c|c|c|c|}
  \hline
  &Positive&Negative&Neutral\\
  \hline
  \multirow{3}{*}{\textbf{Reference}}& yes it does & no , not & i have not \\
  & yes it is &no it is&hope that helps\\
  &yes it will&no it does&do n't think \\
  \hline
  \multirow{3}{*}{\textbf{RAGE}}&that has a&it does have&do n't know \\
  &it fits perfectly&, it does&n't know the \\
  &am not familiar&but it does&know the answer \\
  \hline
  \multirow{3}{*}{\textbf{SAHSSC-Q}}&yes , it&does not have&know about the \\
  &work with any&not have a&not sure what \\
  & will work with&does n't have&but i have\\
  \hline
  \multirow{3}{*}{\textbf{OAAG}}&yes , it &it does not  &is a little\\
  &yes it does&does not have& hope this helps\\
  & yes it will&not have a& had any problems\\
  \hline
  \end{tabular}

\label{trigram}
\end{table}

\subsubsection{\textbf{Repetition Analysis in Answers}}
\begin{figure}
\centering
\includegraphics[width=0.48\textwidth]{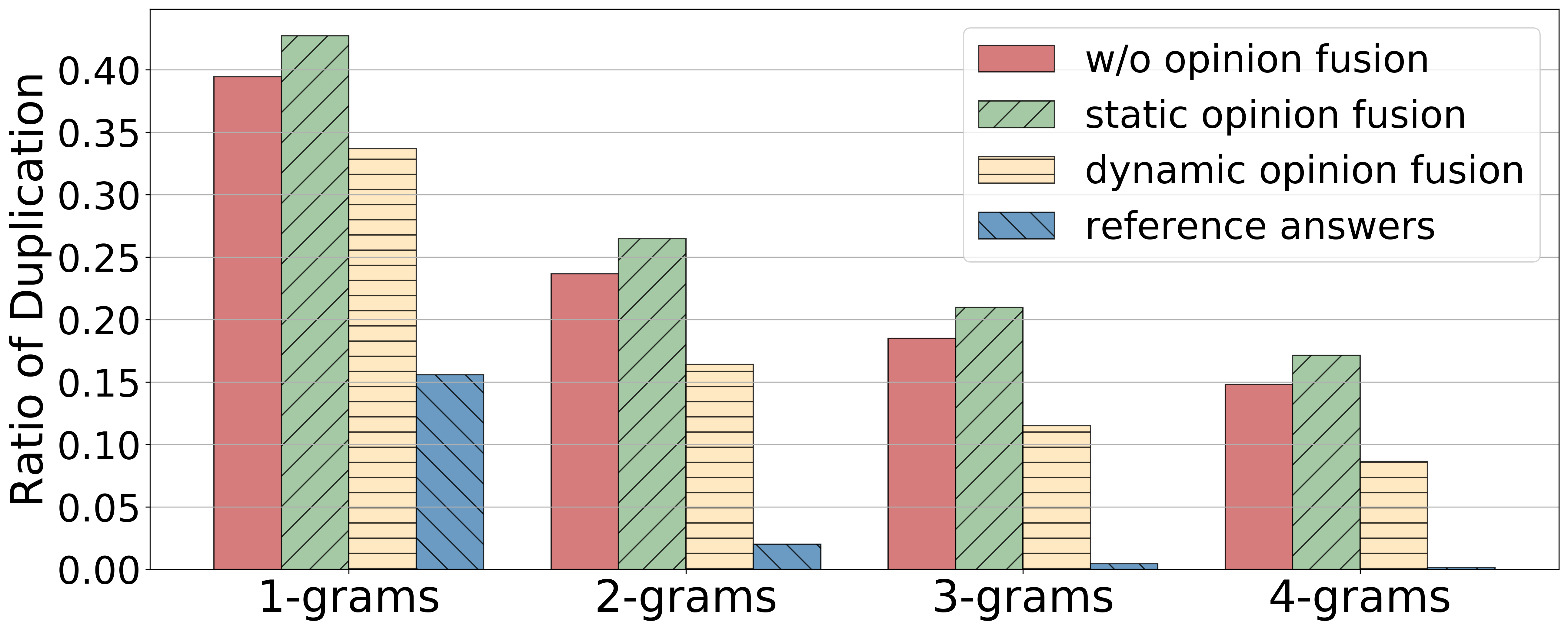}
\caption{Repetition analysis in generated answers}
\label{dup}
\end{figure}

Finally, to empirically answer the research question \textbf{RQ5}, we investigate the diversity of the generated answers by different variants of OAAG. We adopt the $1 - \text{Distinct-N}$ as the evaluation metric to report the ratio of n-grams duplication. Figure~\ref{dup} summarizes the results of the ground-truth answers and the generated answers with static and dynamic opinion fusion or without opinion fusion. We observe that although static opinion fusion improves the performance in content preservation, it causes repetition issues when generating answers since the static review-level attention attends the most influential review during the whole generation process. Take the third case in Table~\ref{case} for instance. OAAG-S may repeatedly emphasize the same thing, e.g., ``it will not fit a carry-on bag" and ``I don't think it will fit a carry-on" in this case. However, dynamic fusion effectively addresses this issue by attending to different opinionated reviews along with the answer decoding procedure, so that it can generate answers with a higher diversity of information.

\begin{table}
\caption{Case study of generated answers}
\centering
  \begin{tabular}{!{\vline width 1pt}p{0.47\textwidth}!{\vline width 1pt}}
  \Xhline{1pt}
  \textbf{Question}: Are the leg height adjustments easy to manipulate and change? (\textit{Electronics})\\
  \textbf{Reference Answer}: Yes they are. Very easy and they stay where you set them. (\textit{positive}) \\
  \hline
  \textbf{S2SAR}: The leg is on the back of the head unit. I don't know about the leg height, but I don't see why it wouldn't work. (\textit{neutral} $\times$)\\
  \textbf{RAGE}: I don't think so. I haven't had any problem with the leg of the camera, but I don't think it would be easy enough. (\textit{negative} $\times$)\\
  \textbf{SAHSSC-Q}: I don't think so . I don't know about the height, but i don't think so. (\textit{negative} $\times$)\\
  \hline
  \textbf{OAAG-S}: Yes, they are very easy to set up. I have used them for several years and haven't had any issues with them. (\textit{positive} $\surd$)\\
  \textbf{OAAG-D}: Yes, you can adjust the leg. If you don't have enough space to mount it to a tripod, you should be fine. (\textit{positive} $\surd$)\\
  \Xhline{1pt}
  \end{tabular}
  
  \begin{tabular}{!{\vline width 1pt}p{0.47\textwidth}!{\vline width 1pt}}
  \Xhline{1pt}
  \textbf{Question}: Can these be used in the microwave? (\textit{Home\&Kitchen})\\
  \textbf{Reference Answer}: Yes, they can. I'm not much of a microwave user, but have used them there on occasion. (\textit{positive}) \\
  \hline
  \textbf{S2SAR}: I have not used them in the microwave, but i don't see why not. I use them in the microwave and have had no problems with them. (\textit{paradoxical} $\times$)\\
  \textbf{RAGE}: I don't see why not. I don't think they would be fine in the microwave, but thank you for your inquiry. (\textit{paradoxical} $\times$)\\
  \textbf{SAHSSC-Q}: I don't see why not. I don't see why you couldn't use them in the microwave. I use them in the microwave and they are microwave safe. (\textit{positive} $\surd$)\\
  \hline
  \textbf{OAAG-S}: Yes, they can be used on any kind of microwave, but I don't think they would withstand high temperatures. I haven't had any problems with them at all. (\textit{positive} $\surd$)\\
  \textbf{OAAG-D}: Yes, they can be used in the microwave. I don't see why not, but they are made of stainless steel so I don't know if they would be microwave safe. (\textit{positive} $\surd$)\\
  \Xhline{1pt}
  \end{tabular}
  
  \begin{tabular}{!{\vline width 1pt}p{0.47\textwidth}!{\vline width 1pt}}
  \Xhline{1pt}
  \textbf{Question}: Will it fit into a carry-on bag? (\textit{Sports\&Outdoors})\\
  \textbf{Reference Answer}: No, I don't believe it will. (\textit{negative}) \\
  \hline
  \textbf{S2SAR}: I don't see why it wouldn't fit in the bag, it would be too big. (\textit{paradoxical} $\times$)\\
  \textbf{RAGE}: I don't see why it wouldn't be a problem. I don't think it would be a problem. (\textit{paradoxical} $\times$)\\
  \textbf{SAHSSC-Q}: I don't think it would fit into a carry-on bag. It doesn't fit into a carry-on bag. (\textit{negative} $\surd$) \\
  \hline
  \textbf{OAAG-S}: No, it will not fit a carry-on bag. I have not tried it on a bag yet, but I don't think it will fit a carry-on. (\textit{negative} $\surd$)\\
  \textbf{OAAG-D}: I don't think it will, it 's too big on the back of the chair. I'm not sure if it's the same size as the carry-on. (\textit{negative} $\surd$)\\
  \Xhline{1pt}
  \end{tabular}

\label{case}
\vspace{-0.3cm}
\end{table}

\subsubsection{\textbf{Case Study}}
Table~\ref{case} shows three specific cases with generated answers by different methods. We observe that general generation methods, i.e., S2SAR and RAGE, generate answers with random opinions, which may contrast to the actual experience from other customers. For instance, in the first case concerning a subjective question, the generated answer by RAGE holds a contradictory opinion with the reference answer, which is also the top-voted answer by other customers, so that it is likely to be an unhelpful answer. Moreover, in the second and third cases concerning some objective issues, S2SAR and RAGE even generate answers with paradoxical opinions, which indicates the necessity of the guidance of certain opinion for answer generation. Besides, although SAHSSC-Q is more sensitive to the opinion due to its consideration of user opinions, the generated answer may not be specific to the question and provide limited information. 

OAAG effectively overcomes these shortcomings and generates opinion-aware and informative answers, which are more valuable for customers. Apparently, the generated answers by OAAG are more straightforward and helpful for the customers to address their concerns. In particular, OAAG-S tends to generate coherent and consistent answers, while OAAG-D can incorporate diverse but relevant information from different reviews into the generated answers. As the 1st and 2nd case, OAAG-D not only gives the direct answer to the given question, but also provides some extra information or explanations. These cases can be served as the evidence for answering the research question \textbf{RQ5}.

\section{Conclusions}
We propose Opinion-Aware Answer Generator to generate opinion-aware answers for review-driven question answering by jointly learning answer generation and opinion mining tasks with opinion fusion. 
In specific, a multi-view pointer generator network with static and dynamic opinion fusion is designed to generate opinion-aware answers for review-driven question answering. 
The experimental results on real-world E-Commerce QA datasets show that our method not only outperforms existing generation methods in content preservation, but also guarantees to generate opinionated and informative answers.

\balance
\bibliographystyle{ACM-Reference-Format}
\bibliography{bibliography}

\end{document}